\documentclass{article}
\usepackage{spconf,amsmath,graphicx}
\usepackage{amsfonts}
\usepackage{algorithmic}
\usepackage{algorithm}
\usepackage{array}
\usepackage[caption=false,font=normalsize,labelfont=sf,textfont=sf]{subfig}
\usepackage{textcomp}
\usepackage{stfloats}
\usepackage{url}
\usepackage{verbatim}
\usepackage{cite}
\usepackage{booktabs}

\newcommand{\ie}{\textit{i}.\textit{e}.}
\newcommand{\eg}{\textit{e}.\textit{g}.}
\usepackage[T1]{fontenc}
\usepackage{multirow}
\usepackage{multicol}

\newenvironment{tightcenter}{%
  \setlength\topsep{3pt}
  \setlength\parskip{3pt}
  \begin{center}
  \begin{minipage}{.42\textwidth}
}{%
  \end{minipage}
  \end{center}
}

\title{A comparative study of perceptual quality metrics for \\ audio-driven talking head videos}
\name{Weixia~Zhang, Chengguang~Zhu, Jingnan~Gao, Yichao~Yan, Guangtao~Zhai, Xiaokang~Yang}
\address{MoE Key Lab of Artificial Intelligence, AI Institute, Shanghai Jiao Tong University}
\begin{document}
%
\maketitle
\begin{abstract}
The rapid advancement of Artificial Intelligence Generated Content (AIGC) technology has propelled audio-driven talking head generation, gaining considerable research attention for practical applications. However, performance evaluation research lags behind the development of talking head generation techniques. Existing literature relies on heuristic quantitative metrics without human validation, hindering accurate progress assessment. To address this gap, we collect talking head videos generated from four generative methods and conduct controlled psychophysical experiments on visual quality, lip-audio synchronization, and head movement naturalness.  Our experiments validate consistency between model predictions and human annotations, identifying metrics that align better with human opinions than widely-used measures. We believe our work will facilitate performance evaluation and model development, providing insights into AIGC in a broader context. Code and data will be made available at \url{https://github.com/zwx8981/ADTH-QA}.
\end{abstract}
\section{Introduction}
\label{sec:intro}
\begin{figure}[tb]
    \centering
    \captionsetup{justification=centering}
    \includegraphics[width=0.49\textwidth]{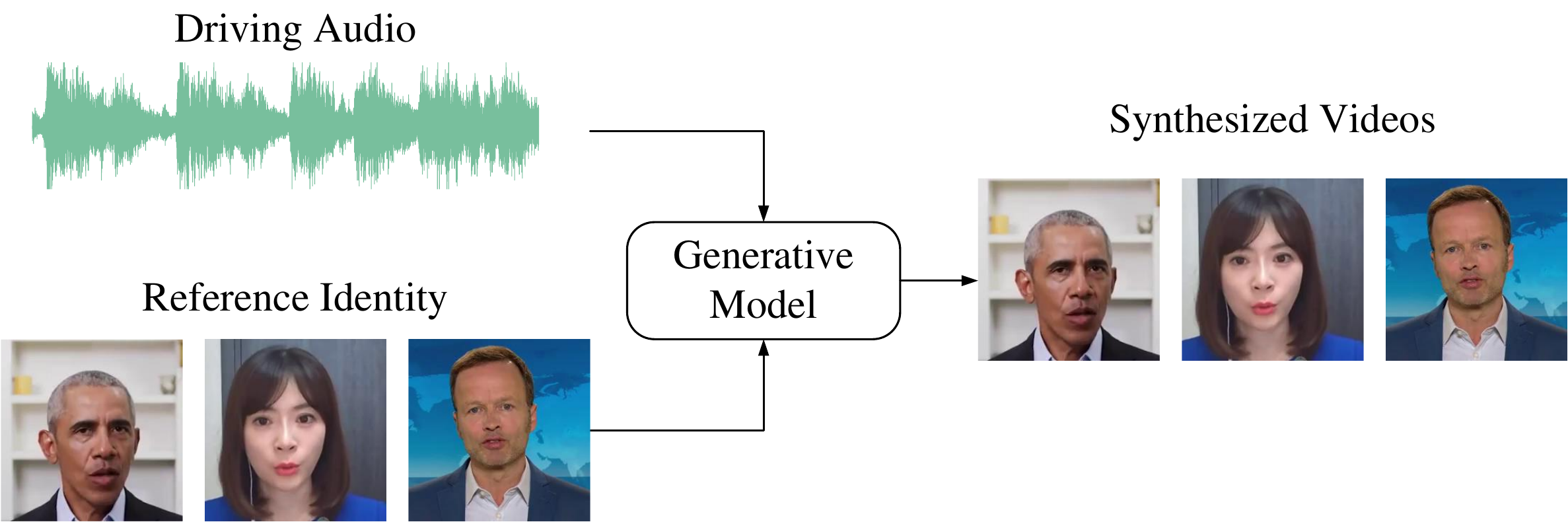}
  \caption{Illustration of audio-driven talking head generation.}
\label{fig:demo}
\end{figure}

The concept of the ``metaverse" can be traced back to \textit{Snow Crash} a science fiction novel published in 1992 that depicts a virtual world accessed by humans through digital avatars for interaction. Inhabiting the metaverse, digital humans play a vital role in shaping this virtual world. The audio-driven talking head generation~\cite{toshpulatov2023talking} stands out among various technologies, which animates faces to match a given audio input as demonstrated in Fig.~\ref{fig:demo}. Recent advancements in this field~\cite{jamaludin2019you, suwajanakorn2017synthesizing, zhou2019talking, chen2018lip} underscore its potential across a wide array of applications, such as newscasting, customer service, gaming, and smart healthcare, etc.

Accurate assessment of research advancements in this field is equally crucial to the development of new technologies. As audio-driven talking head generation is inherently multimodal, its evaluation requires consideration of various output aspects. In existing research, visual fidelity metrics, such as the Peak Signal-to-Noise Ratio (PSNR) and Structural Similarity Index (SSIM)~\cite{wang2004image}, are commonly utilized to assess the perceptual quality of generated talking head videos. Animated talking heads are expected to display synchronized lip motions with audio and deliver natural head movements. SyncNet model~\cite{chung2017out} has become a popular tool for assessing lip-audio synchronization in the literature. Existing works use the average Euclidean landmark distance (LMD) between generated and reference videos~\cite{chen2018lip} as a proxy metric for assessing head movement naturalness~\cite{zhou2020makelttalk}. We ask the key question:
\begin{tightcenter}
    \textit{Can we trust these metrics as surrogates for humans?}
\end{tightcenter}

The prevalent evaluation methods for audio-driven talking head generation exhibit many limitations. Diverse data sources used in training and testing hinder performance comparability across studies. Given the ill-posed nature of the problem, where one audio input can lead to multiple valid outputs, the effectiveness of fidelity-based metrics like PSNR and SSIM is questionable. Moreover, the lack of human quality annotations challenges the validation of performance metrics, complicating the reliable evaluation of models and the development of more effective metrics.

In this study, we collect audio-driven talking head videos from four generative methods and conduct psychophysical experiments in a well-controlled lab setting. Participants rate their preferences between video pairs generated by different methods, focusing on visual quality, lip-audio synchronization, and head movement naturalness. We then extensively test various objective metrics against these human judgments to evaluate their effectiveness. Compared with the heuristic performance metric selection scheme, our solution serves as a more rigorous strategy to identify metrics and design choices that better align with subjective perceptions.

In summary, our primary contributions are threefold.
\begin{itemize}
\setlength{\itemsep}{0pt}
\setlength{\parsep}{0pt}
\setlength{\parskip}{0pt}
    \item We assembled a diverse dataset of photorealistic human videos, encompassing variations in gender, age, language, and context, which serves well for the training and evaluation of audio-driven talking head generation models.
    \item We conducted psychophysical experiments to collect human preferences for talking head video pairs, enabling quantitative assessment of objective metrics.
    \item We performed extensive experiments to examine and identify computational models that better align with human judgments.
\end{itemize}

\section{Subjective Study}~\label{sec:subjective}
In this section, we summarize our efforts toward constructing a dataset consisting of audio-driven talking head videos generated by four representative generative methods.
\subsection{Real Human Video Collection}\label{subsec:real_human}
\begin{figure*}[htbp]
    \centering
    \captionsetup{justification=centering}
    \subfloat[French]{\includegraphics[width=0.19\textwidth]{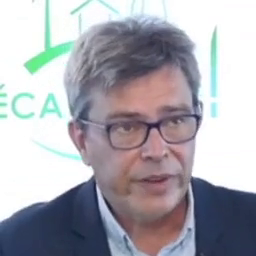}}\hskip.1em
    \subfloat[German]{\includegraphics[width=0.19\textwidth]{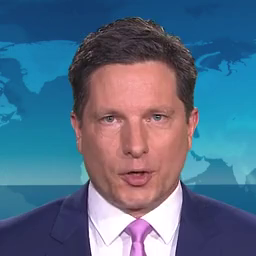}}\hskip.1em
    \subfloat[German]{\includegraphics[width=0.19\textwidth]{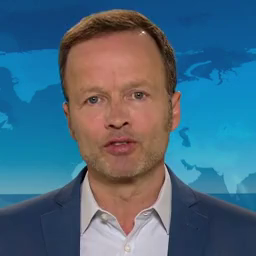}}\hskip.1em
    \subfloat[Mandarin]{\includegraphics[width=0.19\textwidth]{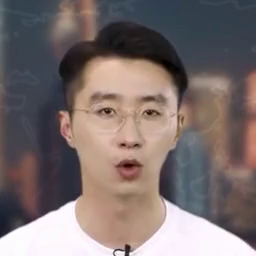}}\hskip.1em
    \subfloat[English]{\includegraphics[width=0.19\textwidth]{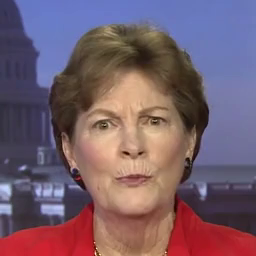}}\hskip.1em
    \subfloat[Japanese]{\includegraphics[width=0.19\textwidth]{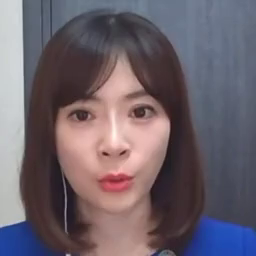}}\hskip.1em
    \subfloat[Mandarin]{\includegraphics[width=0.19\textwidth]{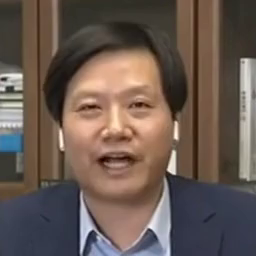}}\hskip.1em
    \subfloat[English]{\includegraphics[width=0.19\textwidth]{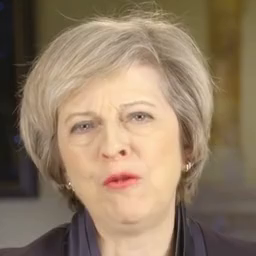}}\hskip.1em
    \subfloat[English]{\includegraphics[width=0.19\textwidth]{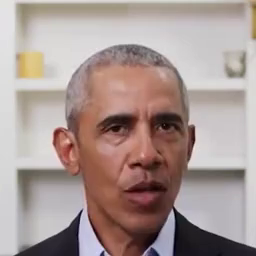}}\hskip.1em
    \subfloat[Spanish]{\includegraphics[width=0.19\textwidth]{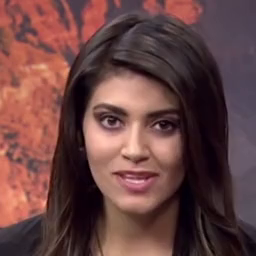}}
  \caption{Thumbnails of real human video sequences in our dataset. The spoken language is marked below each thumbnail.}
\label{fig:real_human}
\end{figure*}

We aim to build a standard playground where different audio-driven talking head generation models can be effectively trained and fairly evaluated. Specifically, we retrieve ten real human videos with paired audio-visual data from sources including HDTF~\cite{zhang2021flow}, \textit{youtube.com}, and \textit{bilibili.com}, which are then processed to highlight the face and upper torso within a uniform frame resolution of $256 \times 256$ pixels at 25 frame-per-second (FPS).  As shown in Fig.~\ref{fig:real_human}, the resulting video set covers both male and female identities of different ages and speaking in six different languages. 

\subsection{Talking Head Videos Generation}\label{subsec:generation}
\begin{figure*}[t]
    \centering
    \captionsetup{justification=centering}
    \subfloat[Mouth cavity]{\includegraphics[width=0.24\textwidth]{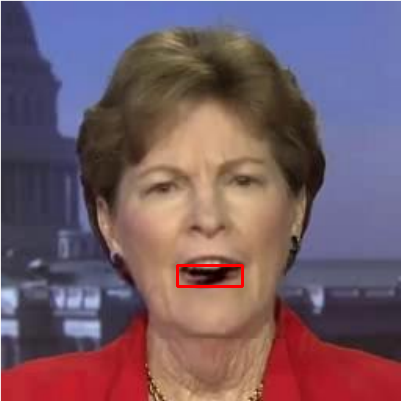}}\hskip0.5em
    \subfloat[Spatial deformation]{\includegraphics[width=0.24\textwidth]{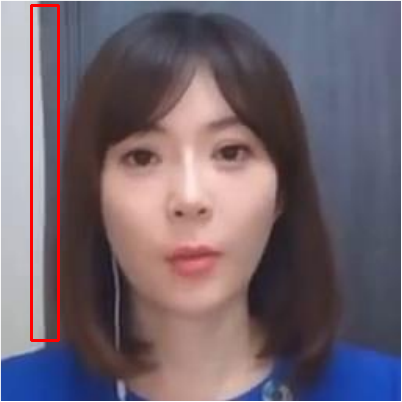}}\hskip0.5em
    \subfloat[Head-torso gap]{\includegraphics[width=0.24\textwidth]{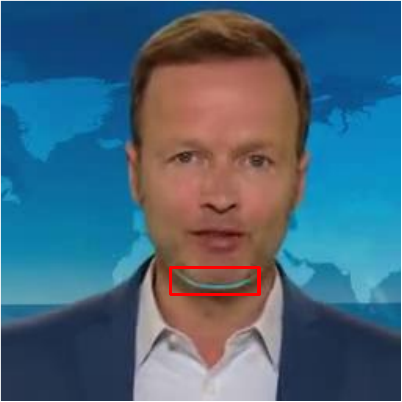}}\hskip0.5em
    \subfloat[Blur face]{\includegraphics[width=0.24\textwidth]{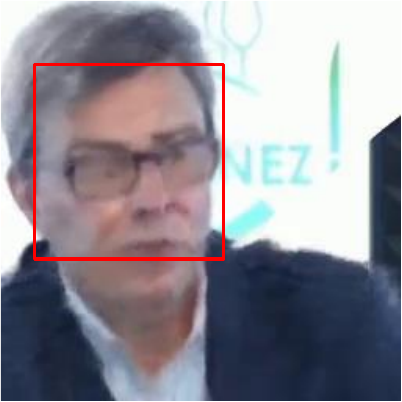}}
  \caption{Example algorithm-related artifacts introduced by (a) Wav2Lip, (b) MakeItTalk, (c) AD-NeRF, and (d) DFA-NeRF. The primary degraded regions are marked with red boxes.}
\label{fig:artifacts}
\end{figure*}

The defects in the generated talking head videos arise from algorithmic imperfections rather than synthesis by explicit model or consequence of acquisition.(see Fig.~\ref{fig:artifacts}). We have $N$ real-human video and audio$\{(\mathbf{v}^{i}, \mathbf{a}^{i})\}_{i=1}^{N}$ corresponding to $N$ different identities, where $\mathbf{v}^{i}=(v_{1}^{i}, ..., v_{T_{v}}^{i})$ and $\mathbf{a}^{i}=(a_{1}^{i}, ..., a_{T_{a}}^{i})$ stand for the $i$-th video and audio sequences with $T_{v}^{i}$ and $T_{a}^{i}$ frames, respectively. A generative model $G_{w}(\cdot)$ parameterized by a vector $w$ either works in an identity-aware or an identity-agnostic manner, where the former separately learn $N$ parameter vectors $\{w_{i}\}_{i=1}^{N}$ to account for each identity, the latter hold a general parameter vector $w$ for any identity. Without loss of generality, we use the first frame of the $i$-th video $v_{1}^{i}$ and $\mathbf{a}^{i}$ as the conditions to the $i$-th output video $\hat{v}^{i}$: $\mathbf{\hat{v}}^{i} = G_{w}(\mathbf{a}^{i}, v_{1}^{i})$ for an identity-agnostic method. Original videos are divided into training and testing segments, using the first $90\%$ of frames for training and the rest for testing. Consequently, the length of test videos ranges from 10 to 15 seconds.  

\begin{figure*}[t]
    \centering
    \captionsetup{justification=centering}
    \subfloat[]{\includegraphics[width=0.42\textwidth]{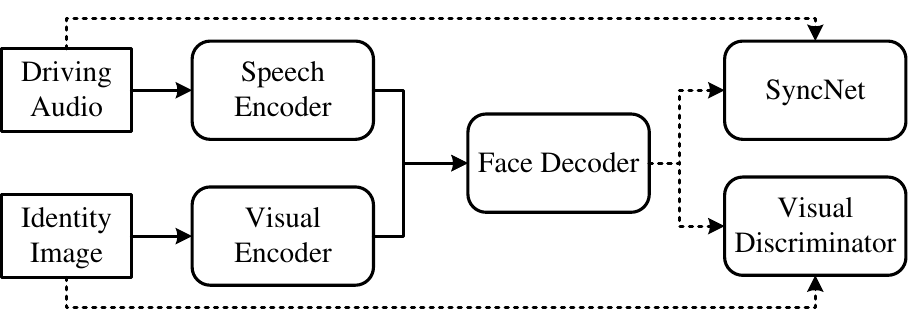}}\hskip2em
    \subfloat[]{\includegraphics[width=0.42\textwidth]{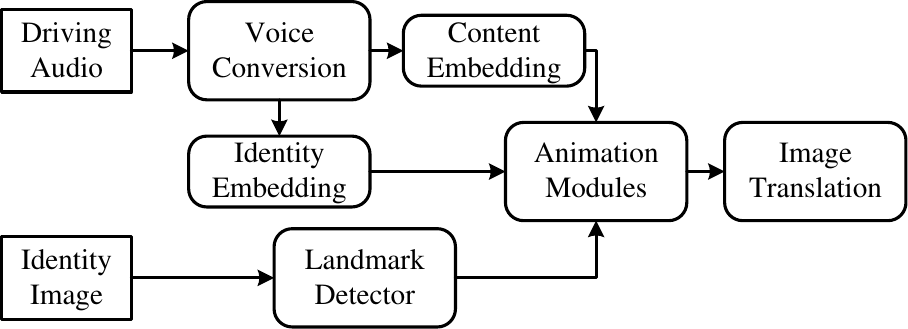}}\hskip2em
    \subfloat[]{\includegraphics[width=0.42\textwidth]{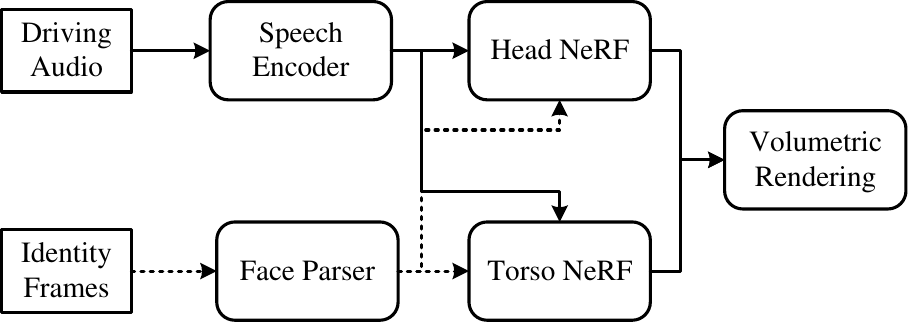}}\hskip2em
    \subfloat[]{\includegraphics[width=0.42\textwidth]{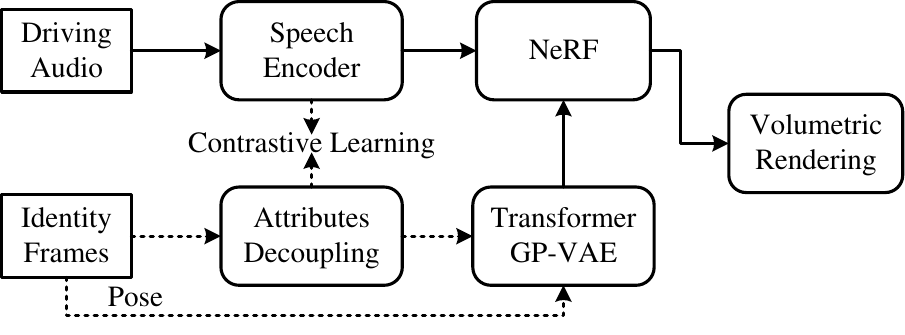}}
  \caption{System diagrams of (a) Wav2Lip, (b) MakeItTalk, (c) AD-NeRF, and (d) DFA-NeRF. Grey arrows correspond to the operations performed in the training phase only.}
\label{fig:four_methods}
\end{figure*}

We select four representative audio-drive talking head generation methods, namely Wav2Lip~\cite{prajwal2020lip} and MakeItTalk~\cite{zhou2020makelttalk} which employ generative adversarial network (GAN), AD-NeRF~\cite{guo2021ad} and DFA-NeRF~\cite{yao2022dfa} which utilize neural radiance fields (NeRF). Wav2Lip and MakeItTalk are identity-agnostic methods, AD-NeRF and DFA-NeRF are identity-aware methods by design. We visualize the system diagram of each method in Fig.~\ref{fig:four_methods} and provide the corresponding descriptions as follows: 

\noindent 1. \textbf{Wav2Lip}~\cite{prajwal2020lip} is a GAN-based method that aims at accurately morphing lip movement of arbitrary identities with the help of a lip-sync discriminator, \ie, SyncNet~\cite{chung2017out}. Meanwhile, it also trains a visual quality discriminator to mitigate undesired blurry effects in morphed regions. 

\noindent 2. \textbf{MakeItTalk}~\cite{zhou2020makelttalk} uses a voice conversion model to disentangle the input audio signal into the content and identity embedding, which controls the motions surrounding the lips and other regions, respectively. It employs a detector to obtain the facial landmarks first, which are then fed into the animation modules. Finally, an image-to-image translation module is used for video synthesis. 

\noindent 3. \textbf{AD-NeRF}~\cite{guo2021ad} extends NeRF by incorporating the audio features as the condition. With the help of a face parser, AD-NeRF trains two individual neural radiance fields to model the dynamics of the head and torso, respectively. Volumetric rendering is utilized to obtain the output videos. We use the original video as the input following the pipeline of the authors' implementation. 

\noindent 4. \textbf{DFA-NeRF}~\cite{yao2022dfa} introduces a contrastive learning scheme to synchronize the audio and lip motions. In addition, it decouples the face attributes into separate features, based on which a Transformer-based variational autoencoder sampled from the Gaussian process is trained to learn natural-looking head poses and eye blinks. The generated features are concatenated and serve as the condition of a NeRF before the final volumetric rendering stage.

\subsection{Psychophysical Experiment}\label{subsec:subjective_study}
We conduct three psychophysical experiments to evaluate human preferences regarding visual quality, lip-audio synchronization, and head movement naturalness in generated talking head videos. Fifteen participants with normal or corrected-to-normal vision are recruited. Utilizing the two-alternative forced choice (2AFC) method, participants compare pairs of videos generated by different methods (\eg, MakeItTalk vs. AD-NeRF) for the same audio and identity, choosing the preferred one based on one of the three criteria. Experiments are conducted on a $24^{''}$  LCD monitor at $1920 \times 1080$ resolution in an office setting with standard lighting (around 200 lux), free from reflective surfaces. Subjects are required to take a break during the experiment to avoid the fatigue effect. Given four audio-driven talking head generation methods, we have $\tbinom{4}{2} = 6$ video pairs for each combination of identity and driving audio. Participants are asked to annotate all pairs, resulting in a total of $2,700$ human annotations.

\subsection{Data Analysis}\label{subsec:analysis}
We count the votes for different audio-driven talking head generation methods and normalize the votes by the total number of human ratings at each session, showing the results in Table~\ref{tab:vote}. We have several useful observations. First, although AD-NeRF harnesses NeRF's high-fidelity rendering capabilities, it produces videos with annoying head-torso gaps (see Fig.~\ref{fig:artifacts} (c)), significantly compromising visual quality due to its separate head and torso training methodology. Second, DFA-NeRF utilizes face attribute disentanglement and the GP-VAE model to achieve high-quality videos with natural head movements. However, it still confronts convergence challenges in training with uncommon poses (as depicted in Figure \ref{fig:artifacts} (d)). Third, equipped with SyncNet~\cite{chung2017out} as a discriminator, Wav2Lip excels in lip-audio synchronization but falls short in animating other facial areas. In contrast, AD-NeRF and DFA-NeRF render videos that display more natural head movements, mostly due to the inherent capacity of NeRF to capture dynamic features. Fourth, participants agreed more on visual quality and head movement naturalness sessions than lip-audio synchronization, indicating no dominant method in the latter.

\begin{table}[t]
  \centering
  \caption{Normalized votes of different methods at different sessions of psychophysical experiments. A higher value indicates more subjects prefer the respective method.}\label{tab:vote}
  \setlength{\tabcolsep}{1.5mm}
  \begin{tabular}{l|cccc}
      \toprule
      {Methods} & Wav2Lip & MakeItTalk & AD-NeRF  & DFA-NeRF\\
     \midrule
    Visual & 0.3144 & 0.2854 & 0.0539 & 0.3462\\
    Lip & 0.3089 & 0.2256 & 0.3133 & 0.1522\\
    Head & 0.0078 & 0.2378 & 0.3600 & 0.3944\\
     \bottomrule
  \end{tabular}
\end{table}

\begin{figure*}[t]
    \centering
    \captionsetup{justification=centering}
    {\includegraphics[width=1\textwidth]{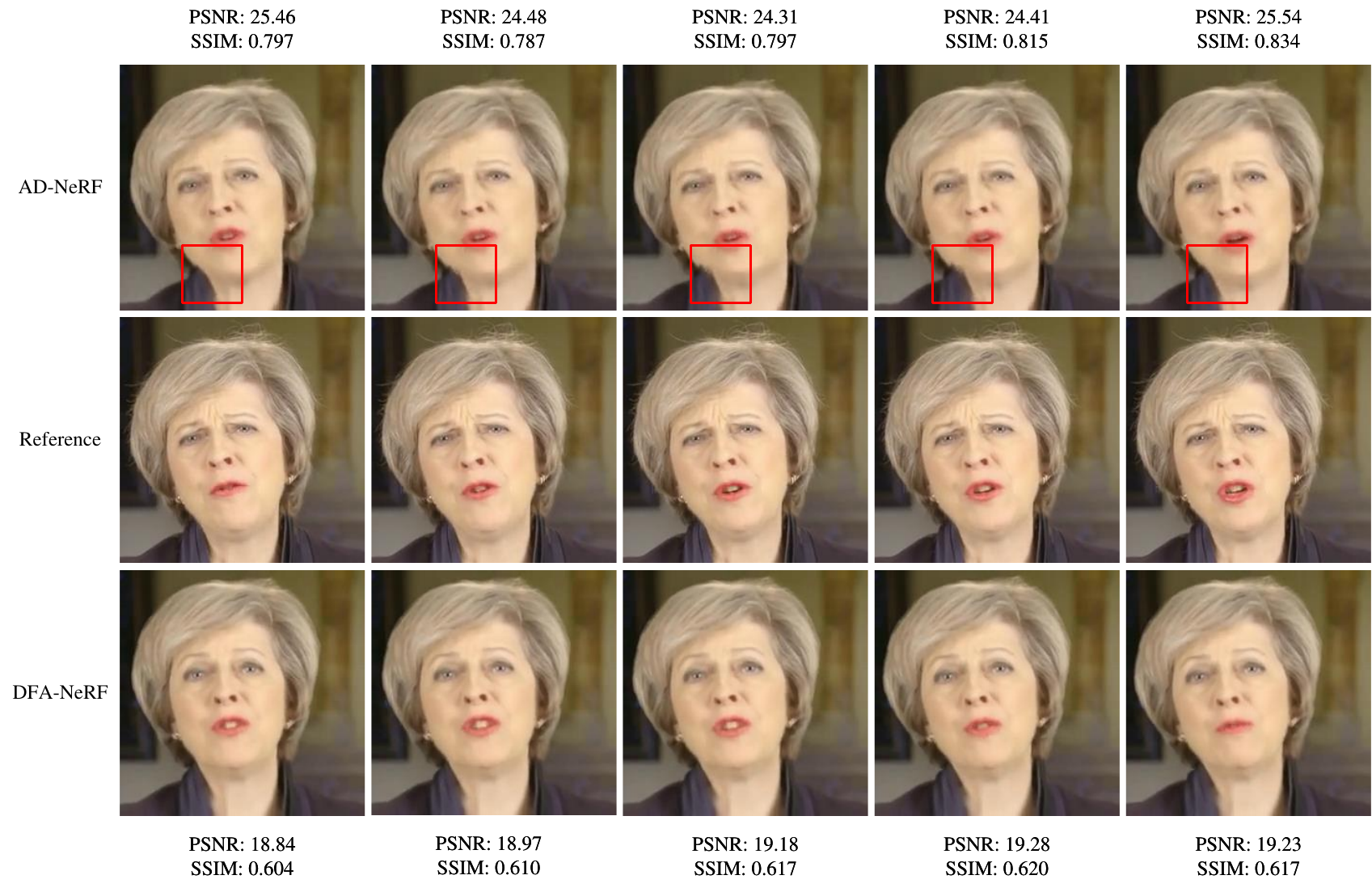}}
  \caption{Consecutive frames of videos generated by AD-NeRF and DFA-NeRF, along with the PSNR and SSIM values with reference to the corresponding reference frames at the same timestamps. Unnatural artifacts are marked in red boxes}
\label{fig:psnr_ssim}
\end{figure*}

\section{Evaluation of Objective Metrics}
\label{sec:objective}
We aim to identify effective quality metrics for audio-driven talking head videos by measuring their alignment with human evaluations. We employ the dataset described in Sec.~\ref{sec:subjective} for testing, following~\cite{zhang2018unreasonable, ding2021comparison} by using the 2AFC score as the performance measure for all objective metrics. Specifically, the 2AFC score is computed by $p \times q + (1 - p) \times (1 - q)$, where $p$ is the percentage of human votes and $q = \{0, 1\}$ is the binary prediction by an objective metric. A higher 2AFC score indicates better alignment. We evaluate objective metrics on video pairs with a clear human preference where one side has a vote rate $p\geq0.75$. Metrics for visual quality, lip-audio synchronization, and head movement naturalness are successively evaluated. 

\subsection{Visual Quality}\label{subsec:render}
In addition to PSNR and SSIM~\cite{wang2004image}, the \textit{de facto} standard visual quality metrics used in the field of audio-driven talking head generation, we evaluate two additional full-reference (FR) image quality metrics: LPIPS~\cite{zhang2018unreasonable}, DISTS~\cite{ding2022image}, five no-reference (NR) image quality models: CPBD~\cite{narvekar2011no}, MUSIQ~\cite{ke2021musiq}, DBCNN~\cite{zhang2020blind}, UNIUQE~\cite{zhang2021uncertainty}, LIQE~\cite{zhang2023blind}, and four video quality assessment methods: MDTVSFA~\cite{li2021unified}, FAST-VQA~\cite{wu2022fast}, DOVER~\cite{wu2022Exploring}, Li22~\cite{li2022blindly}. While a video quality model directly estimates the perceptual quality of $\mathbf{v}$ as $q_{v} = \hat{q}(\mathbf{v})$, an image quality model generally works with every single frame $v_{i} \in \mathbb{R}^{N}$ first and then convert the frame-level quality predictions to a video-level score, either in no-reference manner $q_{v} = T(\{\hat{q}(v_{i})\}_{i=1}^{T_{v}})$ or full-reference manner $q_{v} = T(\{\hat{q}(v_{i}, r_{i})\}_{i=1}^{T_{v}})$, where $T(\cdot)$ denotes a temporal pooling method, and $r_{i}$ is the $i$-th frame of the corresponding reference video $\mathbf{r} \in \mathbb{R}^{N \times T_{v}}$. We experiment with three different temporal pooling methods for aggregating frame-level predictions from image quality models: mean pooling, hysteresis pooling, and percentile pooling~\cite{tu2020comparative}. 

We summarize the results in Table~\ref{tab:2afc_vqa}, from which we have several interesting observations. \textbf{First}, the standard metrics PSNR and SSIM correlate poorly with human judgment, primarily due to their strict spatial registration requirement. These metrics excessively penalize generated frames that viewers perceive as reasonably quality but lack precise spatial alignment with reference frames. To illustrate this issue more intuitively, we compare frames produced by AD-NeRF~\cite{guo2021ad} and DFA-NeRF~\cite{yao2022dfa} alongside their reference video in Fig.~\ref{fig:psnr_ssim}. Despite PSNR and SSIM metrics favoring AD-NeRF, subjects show a preference for DFA-NeRF's video, which is artifact-free compared with that of AD-NeRF's video. Being robust to geometric distortion and textural re-sampling by design, DISTS~\cite{ding2022image} exhibits a relatively higher tolerance to spatial misalignment. \textbf{Second}, without accessing the reference video, some NR methods outperform the top-performing FR method DISTS by large margins, indicating the promise of NR metrics in handling this ill-posed problem. \textbf{Third}, image quality models generally work better with percentile pooling, implying that the overall visual quality of a talking head video is dominated by the lowest quality dominate the global video quality. \textbf{Last}, adopting a mixed-dataset training strategy, MDTVSFA~\cite{li2021unified} and Li22~\cite{li2022blindly} present higher performance. Turning our gazes towards the NR methods, a similar trend emerges with two competitive models, UNIQUE~\cite{zhang2021uncertainty} and LIQE~\cite{zhang2023blind}, also being trained on the combination of multiple datasets. 
This suggests that leveraging diverse source domains for training tends to enhance model transferability.

\begin{table}[t]
  \centering
  \caption{2AFC scores for visual quality metrics.}\label{tab:2afc_vqa}
  \begin{tabular}{l|ccc}
      \toprule
     {Metrics} & Mean & Percentile & Hysteresis \\
    \hline
        PSNR & 0.4229 & 0.3771 & 0.4229\\
        SSIM & 0.4229 & 0.3771 & 0.4438 \\
        LPIPS & 0.4476 & 0.5010 & 0.4476\\
        DISTS & 0.5943 & 0.5981 & 0.5943 \\
    \midrule
        CPBD  & 0.5981 & 0.5981 & 0.6229 \\
        MUSIQ  & 0.6648 & 0.7429 & 0.6686 \\
        DBCNN & 0.5943 & 0.5657 & 0.5943 \\
        UNIQUE & 0.6476 & 0.7848 & 0.6686 \\
        LIQE & 0.7048 & 0.7258 & 0.7048 \\
    \midrule
        MDTVSFA & \multicolumn{3}{c}{0.6950}\\
        FAST-VQA & \multicolumn{3}{c}{0.5457}\\
        DOVER & \multicolumn{3}{c}{0.5795}\\
        Li22 & \multicolumn{3}{c}{0.7751}\\
    \bottomrule
   \end{tabular}
\end{table}

\begin{table}[t]
  \centering
  \caption{2AFC scores for lip-audio synchronization.}\label{tab:2afc_lip}
  \begin{tabular}{ccccc}
      \toprule
      LMD${_l}$ & LSE-D & LSE-C & LSE-O & SparseSync\\
           \hline
     0.7556 & 0.2815 & 0.2333 & 0.5815  & 0.6778\\
     \bottomrule
  \end{tabular}
\end{table}

\begin{table}[t]
  \centering
  \caption{2AFC scores for head movement naturalness. }\label{tab:2afc_head}
  \begin{tabular}{cc}
      \toprule
      LMD${_h}$ & ViSiL \\
      \hline
        0.6365 & 0.8492\\
     \bottomrule
  \end{tabular}
\end{table}

\subsection{Lip-audio Synchronization}\label{subsec:lip}
We compare five lip-audio synchronization evaluation metrics: LMD$_{l}$, LSE-D, LSE-C, LSE-O, and SparseSync~\cite{Iashin2022sparse} in Table~\ref{tab:2afc_lip}. 
LMD$_{l}$ computes the average Euclidean distance between all facial landmark locations around the mouth region of the generated video $\mathbf{v}$ and that of the reference video $\mathbf{r}$ as: $\mathrm{LMD}_{l} = \frac{1}{M} \sum_{i=1}^{M} ||f_{d}^{i}(\mathbf{v}),  f_{d}^{i}(\mathbf{r})||_{2}$, where $f_{d}(\cdot)$ denotes a facial landmark detector, and the locations of $M$ pre-defined facial landmarks around the mouth region\footnote{We adopt the 68 points facial landmark definition by Dlib, among which 12 landmarks correspond to the mouth region.} in total are indexed by $i$. Following the notation of~\cite{prajwal2020lip}, LSE-D, LSE-C, and LSE-O are based on SyncNet~\cite{chung2017out}, corresponding to the average Euclidean distance between visual and audio embeddings, the lip-audio synchronization confidence, and the time offset according to a pre-defined temporal grid, respectively. SparseSync~\cite{Iashin2022sparse} fuses audio and visual information from pre-trained sound classification and action recognition models and predicts the time offset over a pre-defined time grid. We gain several useful observations. First, by measuring the distance of landmarks around the mouth region between the generated and reference video, LMD achieves the best result. Second, among three SyncNet-based metrics, only LSE-O displays moderate effectiveness, casting doubt on their widespread application as quantitative lip-audio synchronization measures in the literature.  Third, being designed to select informative audio-visual information to estimate the audio-visual temporal offset more precisely, SparseSync~\cite{Iashin2022sparse} outperforms LSE-O. In addition, despite the impressive performance of LMD$_{l}$,  it requires access to the reference video, restricting its application when no reference information is available. Under such circumstances, SparseSync acts as an appropriate substitute for quantitatively evaluating lip-audio synchronization. 

\subsection{Head Movement Naturalness}\label{subsec:head}
While the talking head movement patterns differ across subjects and conversations, defining the \textit{naturalness} of head movement unambiguously remains a highly nontrivial task. Similar to LMD${_l}$, we compute the head movement similarity between the generated video and the real video, yet focusing on the face contour instead, termed as LMD${_h}$. We also present a novel solution from the viewpoint of video-to-video retrieval. Specifically, we employ ViSiL~\cite{kordopatis2019visil} to estimate the spatiotemporal similarity between a generated video and the reference video in a high-level semantic latent space. In contrast to LMD${_h}$ which assumes a generated video to follow precisely the same trajectory as the original video, ViSiL evaluates head movement similarity in a high-level semantic latent space. As shown in Table~\ref{tab:2afc_head}, ViSiL outperforms LMD${_h}$ by a clear margin, which not only verifies the effectiveness of our new solution but also highlights the ill-posed nature of talking head generation.

\subsection{Take-Home Messages}\label{subsec:take_home}
Finally, we arrive at a takeaway list as follows:

\noindent 1. Modern image and video quality models align better with human judgments than classical metrics like PSNR and SSIM, prioritizing their use in this field.

\noindent 2. Prominent artifacts in short segments strongly impact the overall quality of a talking head video, favoring the combination of IQA models and the percentile pooling method.

\noindent 3. Learning a quality metric from multiple source domains is a good practice to achieve quality predictions aligned with humans.

\noindent 4. SyncNet-based metrics may not align well with human judgments, prompting a re-evaluation of their effectiveness in evaluating lip-audio synchronization in talking head videos.

\noindent 5. In contrast to the strong correlation between the driving audio and lip movements, movements in other head areas vary considerably, challenging the use of LMD-based metrics.

\noindent 6. ViSiL is qualified for evaluating the naturalness of head movements in generated talking head videos.

\section{Conclusion}
\label{sec:conclusion}
We have created a dataset with videos generated using four audio-driven talking head methods. Through well-controlled psychophysical experiments, we have collected human preferences on generated videos from three Orthogonal perceptual quality dimensions. Specifically, participants in each session are required to express their preferences for a video pair in terms of visual quality, lip-audio synchronization, and head movement naturalness. Through extensive experiments, we spot objective metrics that align better with human judgments than the heuristically employed ones. We believe our work will shed light on both the evaluation and development of audio-driven talking head generation.



\bibliographystyle{IEEEbib}
\bibliography{Weixia}

\end{document}